%% file: main.tex
\pdfoutput=1
\documentclass{article}

\input{preamble.tex}
\input{preamble/preamble_math}
\input{preamble/definitions_basic}

\crefformat{equation}{(#2#1#3)}
\crefformat{figure}{Figure~#2#1#3}
\crefname{example}{Example}{Examples}
\crefname{lemma}{Lemma}{Lemmas}
\crefname{cor}{Corollary}{Corollaries}
\crefname{theorem}{Theorem}{Theorems}
\crefname{assumption}{Assumption}{Assumptions}

\usepackage{enumitem} %
\usepackage[separate-uncertainty=true,multi-part-units=single]{siunitx} %

\usepackage{upgreek}

\declaretheoremstyle[
spacebelow=\parsep,
    spaceabove=\parsep,
  mdframed={
    backgroundcolor=gray!10!white,     %
    hidealllines=true, 
    innertopmargin=8pt, 
    innerbottommargin=4pt, 
    skipabove=8pt,
    skipbelow=10pt,
    nobreak=true
}
]{grayboxed}

\crefname{gassumption}{Assumption}{Assumptions}

\usepackage{thm-restate}

\usepackage{xcolor}
\input{preamble/commenting.tex}

\usepackage[affil-it]{authblk}

\title{Veridical Data Science for\\ Medical Foundation Models}
\date{}
\author[1,2]{Ahmed Alaa}
\author[1]{Bin Yu}
\affil[1]{University of California, Berkeley}
\affil[2]{University of California, San Francisco}

\begin{document}
\maketitle

\begin{abstract}
The advent of foundation models (FMs) such as large language models (LLMs) has led to a cultural shift in data science, both in medicine and beyond. This shift involves moving away from specialized predictive models trained for specific, well-defined domain questions to generalist FMs pre-trained on vast amounts of unstructured data, which can then be adapted to various clinical tasks and questions. As a result, the standard data science workflow in medicine has been fundamentally altered; the foundation model lifecycle (FMLC) now includes distinct upstream and downstream processes, in which computational resources, model and data access, and decision-making power are distributed among multiple stakeholders. At their core, FMs are fundamentally statistical models, and this new workflow challenges the principles of ``Veridical Data Science'' (VDS) \cite{yu2020veridical,yu2024veridical}, hindering the rigorous statistical analysis expected in transparent and scientifically reproducible data science practices. We critically examine the medical FMLC in light of the core principles of VDS: predictability, computability, and stability (PCS), and explain how it deviates from the standard data science workflow. Finally, we propose recommendations for a reimagined medical FMLC that expands and refines the PCS principles for VDS including considering the computational and accessibility constraints inherent to FMs.
\end{abstract}

Clinical data science combines statistics, computing, and algorithms with domain expertise to extract medical knowledge from data. The traditional data science life cycle (DSLC)~typically begins with a clearly defined domain question—such as a specific prediction task related to a particular clinical outcome in a defined patient cohort—and follows a structured sequence of steps, including data collection, processing, modeling, and interpretation (although there are in practice many loops within and between the steps). However, recent foundation models (FMs) have caused a shift in clinical data science. Unlike the DSLC, the foundation model life cycle (FMLC) does not start with a specific domain question, but aims at training general-purpose models. It uses broad, unstructured hospital and other non-domain data for model pretraining, with the upstream process often inaccessible by the downstream process and users. This paradigm shift marks a transition from specialist to generalist models, from predefined tasks to emergent capabilities, from curated domain-specific datasets to unstructured electronic health records (EHRs), and from purely predictive to generative modeling. Examples of medical FMs include clinical large language models (LLMs) and conversational vision-language models, which users can apply to or fine-tune for specialized problems that differ from the data or tasks on which they were initially pre-trained \cite{saab2024capabilities}.

\begin{figure}[t]
\centering
\includegraphics[width=4.25in]{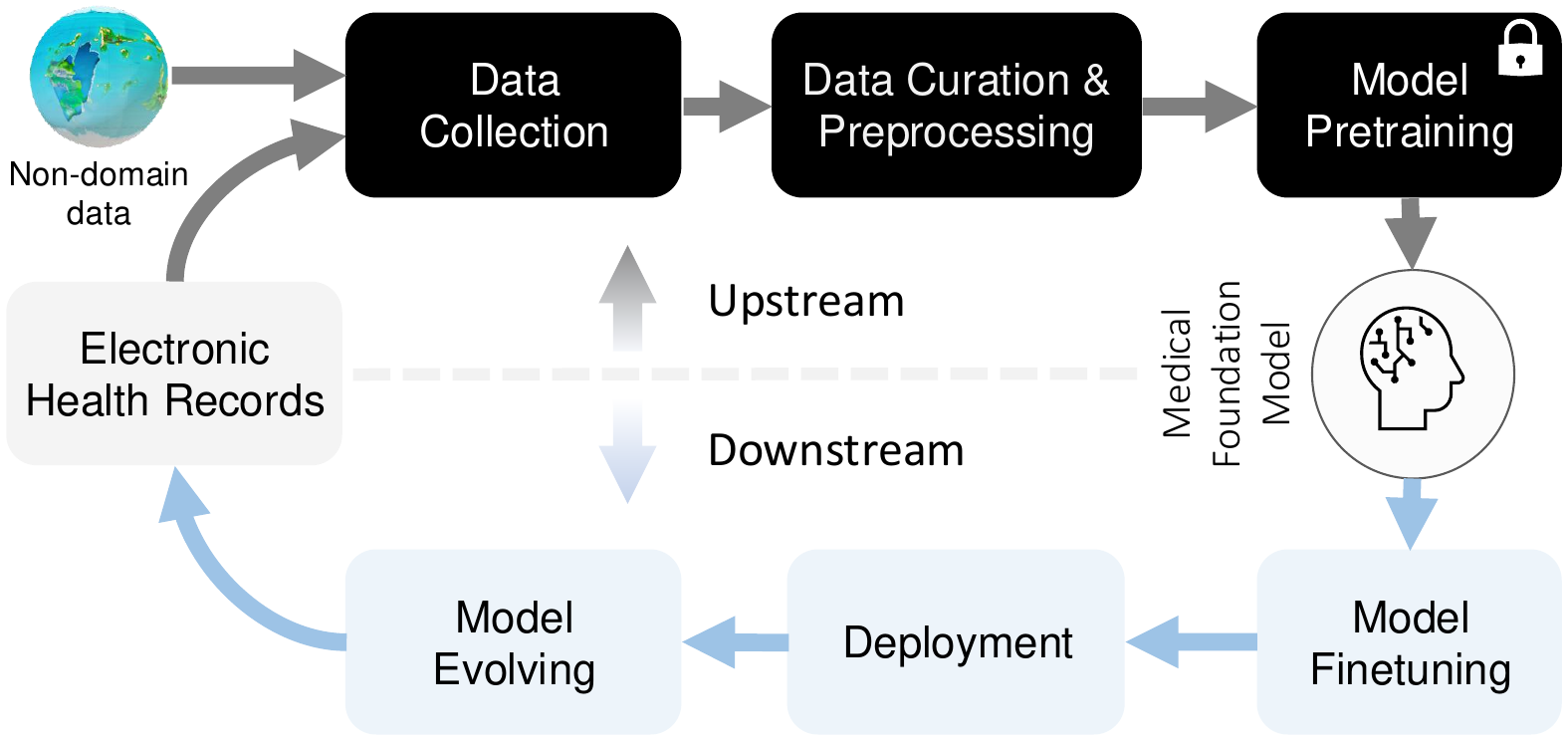}
\caption{{\footnotesize Illustration of the medical foundation model life cycle.}}
\label{Fig1}
\vspace{-.05in}
\rule{\linewidth}{0.45pt}
\vspace{-0.3in}
\end{figure}

The shift from the DSLC to the FMLC brings an expanded scale and scope of underlying operations—it involves distinct upstream and downstream processes with distributed data assets, computational resources, and varying degrees of access among stakeholders. The resulting FM is often deployed as a proprietary “software” that is continuously updated over time (Figure 1), providing users with limited API access without transparency into internal parameters, data used for pretraining and data pre-processing steps. As a result, the FMLC includes numerous (reasonable) human judgment calls that are not transparently documented or communicated across stakeholders. Any variations in these decisions can naturally induce often unacceptable variability in model outputs as shown in recent papers from social science and ecology. These additional uncertainty sources from human judgment calls in the FMLC are not accounted for in standard statistical confidence intervals or hypothesis testing methods, raising serious concerns about the suitability of FMs for inductive inference in scientific and medical applications, where the ultimate goal is to derive reliable, reproducible, and transparent knowledge from data.

A systematic way to consider the statistical implications of the medical FMLC is through the predictability, computability, and stability (PCS) framework and documentation for veridical data science (VDS), introduced by Bin Yu and Karl Kumbier in 2020 for the traditional DSLC \cite{yu2020veridical,yu2024veridical}. Integrating the ``two cultures'' of Breiman (2001) into one \cite{breiman2001statistical}, the PCS framework (including documentation) was originally proposed as a systematic approach to evaluate the impact of human judgment calls on the reliability, reproducibility, and transparency of modeling in the conventional DSLC. It is grounded in three core scientific principles that combine statistics and ML in traditional data science practices: Predictability is the main objective of supervised learning in standard clinical predictive models, providing a ``reality check''—a model can be rejected or revised if it fails to predict new observations in held-out test data. In the 2024 VDS book by Yu and Barter \cite{yu2024veridical}, Predictability has been expanded to be a stand-in for general reality-check including unsupervised learning. Computability concerns computational aspects of the DSLC and includes efficient computing and data-inspired simulations. Stability expands traditional statistical uncertainty considerations (e.g. cross-validation, bootstrapping) to include variability from judgment calls across the DSLC, including those in data-cleaning and algorithm choices. Previous applications of the PCS framework in data science have demonstrated that, in various biomedical contexts, variations induced by perturbing human judgment calls in the DSLC may be comparable to those from bootstrapping a cleaned copy of the (training or test) data. The PCS framework offers a structured approach to address aspects of the FMLC that hinder transparency, reproducibility, and rigor in FM-based data science. Next, we discuss the challenges and recommendations along the three pillars of predictability, computability and stability for FM development to be PCS-compliant during processes of the opaque upstream (e.g. data selection/curation, pretraining) and downstream (e.g. prompting and fine-tuning).

The concept of ``predictability'' as a reality check for FMs raises important questions. Traditional predictive models are developed to solve a well-defined prediction problem, where standard prediction performance measures (including accuracy, sensitivity, and specificity, and for subgroups) on held-out test data may suffice as a ``reality-check''. However, FMs are often used in diverse and non-traditional tasks, such as clinical text summarization. In these contexts, defining a suitable reality check can be nuanced and requires an in-depth understanding of clinical workflows \cite{goodman2024ai}. Despite the availability of public medical benchmarks for predictive tasks, more comprehensive benchmarks are needed to cover both upstream and downstream processes, reflect the dynamic nature of clinical practice, as well as evaluate FM utility in non-traditional tasks such as medical text summarization, automatic clinical note generation or conversational models. For the upstream, basic FDA-like disclosures are recommended on the data, algorithms, prompt-design, their documentation, and release criteria used for FM developers to ensure some essential trust from the downstream process and users, even for their economic gains down the line when paid FMs become common. For both the upstream and downstream, it is recommended that FM developers stress-test pretrained and fine-tuned FMs for high-stakes medical tasks by developing and improving continuously a collection of corner cases (e.g. using medical vignettes) and that they engage appropriate academic and citizen researchers to red-team new FMs for PCS-compliance before release and continuously monitor them over time. 

``Computability'' is another dimension where FMLC differs remarkably from the DSLC. Upstream stakeholders (e.g. tech companies) typically have access to far greater GPU resources than downstream users (e.g. data scientists at hospitals). Retraining models to implement statistical tests by perturbing judgment calls can be prohibitively expensive for downstream users. Initiatives such as the National AI Research Resource pilot program launched by NSF may help bridge the computational gap between upstream and downstream stakeholders in health systems. However, further efforts (e.g. efficient compute advances in algorithms and hardware) are needed to guarantee adequate availability of HIPAA-compliant computational resources to downstream users, ensuring a systematic evaluation process for new releases of FMs. Data-inspired and well-vetted medical simulation models fall also under ``Computability'' and can provide surrogates for reality-check for FMs in certain clinical settings.

Predictability and computability are intertwined with the third pillar: stability. FMLC's stability can be assessed by upstream stakeholders through systematically perturbing and documenting each human judgment call and its impact on performance metrics within computational constraints. This ability to evaluate a model's stability is essential for conducting formal statistical tests on the ``scientific null hypothesis'' of an FM—that the FM does not significantly improve patient outcomes or inform clinical decisions. Pretraining FMs involves numerous upstream engineering decisions that can greatly affect model performance, while downstream users often customize ``prompts'' for specific outputs. Moving forward, it is crucial for upstream stakeholders to document and communicate their judgment calls to downstream users, including details on the data used for pretraining, data preprocessing, optimization algorithms, and hyperparameter tuning. Downstream users should also consider and report the impact of prompt engineering on output variability. Failing to comply with such PCS guidelines may lead to poor scientific reproducibility and compromise the validity of findings based on FMs.

\newpage
\bibliographystyle{unsrt}
\bibliography{refs}

\end{document}

%% file: preamble.tex
\usepackage[utf8]{inputenc} 
\usepackage[T1]{fontenc}    

\usepackage[bitstream-charter]{mathdesign}
\usepackage{amsmath}
\usepackage[scaled=0.92]{PTSans}

\usepackage[
  paper  = letterpaper,
  left   = 1.65in,
  right  = 1.65in,
  top    = 1.0in,
  bottom = 1.0in,
  ]{geometry}

\usepackage[usenames,dvipsnames,table]{xcolor}
\definecolor{shadecolor}{gray}{0.9}

\usepackage[final,expansion=alltext]{microtype}
\usepackage[english]{babel}
\usepackage[parfill]{parskip}
\usepackage{afterpage}
\usepackage{framed}

{\endMakeFramed}

\usepackage{lineno}

\usepackage{ragged2e}

\newcounter{parcount}

\usepackage{graphicx}
\usepackage{wrapfig}
\usepackage[labelfont=bf,font=footnotesize,width=.9\textwidth]{caption}
\usepackage[format=hang]{subcaption}

\usepackage{booktabs,multirow,multicol}       

\usepackage[algoruled]{algorithm2e}
\usepackage{listings}
\usepackage{fancyvrb}
\fvset{fontsize=\normalsize}

\usepackage[colorlinks,linktoc=all]{hyperref}
\usepackage[all]{hypcap}
\hypersetup{citecolor=MidnightBlue}
\hypersetup{linkcolor=MidnightBlue}
\hypersetup{urlcolor=MidnightBlue}

\usepackage[nameinlink]{cleveref}

\lstdefinestyle{mystyle}{
    commentstyle=\color{OliveGreen},
    keywordstyle=\color{BurntOrange},
    numberstyle=\tiny\color{black!60},
    stringstyle=\color{MidnightBlue},
    basicstyle=\ttfamily,
    breakatwhitespace=false,
    breaklines=true,
    captionpos=b,
    keepspaces=true,
    numbers=left,
    numbersep=5pt,
    showspaces=false,
    showstringspaces=false,
    showtabs=false,
    tabsize=2
}

%% file: preamble/preamble_math.tex
\usepackage{centernot}
\usepackage{amsthm}
\usepackage{amsfonts}       
\usepackage{nicefrac}       
\usepackage{mathtools}
\usepackage{amsbsy}
\usepackage{amstext}
\usepackage{amsthm}
\usepackage{thmtools}
\usepackage{thm-restate}

\begingroup
    \makeatletter
    \@for\theoremstyle:=definition,remark,plain\do{%
        \expandafter\g@addto@macro\csname th@\theoremstyle\endcsname{%
            \addtolength\thm@preskip\parskip
            }%
        }
\endgroup

\crefname{lemma}{lemma}{lemmas}
\Crefname{lemma}{Lemma}{Lemmas}
\crefname{thm}{theorem}{theorems}
\Crefname{thm}{Theorem}{Theorems}
\crefname{prop}{proposition}{propositions}
\Crefname{prop}{Proposition}{Propositions}
\crefname{assumption}{assumption}{assumptions}
\crefname{assumption}{Assumption}{Assumptions}

\usepackage{booktabs,arydshln}
\makeatletter
\def\adl@drawiv#1#2#3{%
        \hskip.5\tabcolsep
        \xleaders#3{#2.5\@tempdimb #1{1}#2.5\@tempdimb}%
                #2\z@ plus1fil minus1fil\relax
        \hskip.5\tabcolsep}
\newcommand{\cdashlinelr}[1]{%
  \noalign{\vskip\aboverulesep
           \global\let\@dashdrawstore\adl@draw
           \global\let\adl@draw\adl@drawiv}
  \cdashline{#1}
  \noalign{\global\let\adl@draw\@dashdrawstore
           \vskip\belowrulesep}}
\makeatother

\renewcommand{\epsilon}{\varepsilon}

\declaretheorem[style=plain,name=Theorem]{theorem}

\declaretheorem[style=definition,sibling=theorem,name=Example]{example}

\newenvironment{example*}
 {\pushQED{\qed}\example}
 {\popQED\endexample}
\numberwithin{equation}{section}

%% file: preamble/definitions_basic.tex
\DeclarePairedDelimiterX\Set[1]{\lbrace}{\rbrace}%
{  #1 }



%% file: preamble/commenting.tex
\definecolor{WowColor}{rgb}{.75,0,.75}
\definecolor{SubtleColor}{rgb}{0,0,.50}

\newcounter{margincounter}